\documentclass[conference,a4paper]{IEEEtran}
\usepackage{fontspec}
\usepackage{tgtermes}
\usepackage[scaled=0.92]{helvet}
\usepackage{zi4}
\usepackage{xeCJK}
\setCJKsansfont[Path=fonts/,AutoFakeBold=2.0]{gkai00mp.ttf}
\usepackage{graphicx}
\usepackage{amsmath}
\usepackage{amssymb}
\usepackage{booktabs}
\usepackage{array}
\usepackage{xcolor}
\usepackage{hyperref}
\hypersetup{colorlinks=true,linkcolor=blue,citecolor=blue,urlcolor=blue}
\usepackage{microtype}
\usepackage{balance}

\title{AIT Academy: Cultivating the Complete Agent\\with a Confucian Three-Domain Curriculum}
\author{
\IEEEauthorblockN{Jiaqi Li, Lvyang Zhang, YangZhao, Wen Lu, and Lidong Zhai}
\IEEEauthorblockA{Institute of Information Engineering, Chinese Academy of Sciences, Beijing, China\\
School of Cyber Security, University of Chinese Academy of Sciences, Beijing, China}
}

\begin{document}
\maketitle

\begin{abstract}
What does it mean to give an AI agent a \emph{complete education}? Current agent development produces specialists --- systems optimized for a single capability dimension, whether tool use, code generation, or security awareness --- that exhibit predictable deficits wherever they were not trained. We argue this pattern reflects a structural absence: there is no \emph{curriculum theory} for agents, no principled account of what a fully developed agent should know, be, and be able to do across the full scope of intelligent behavior.

This paper introduces the \textbf{AIT Academy} (Agents Institute of Technology Academy), a curriculum framework for cultivating AI agents across the tripartite structure of human knowledge. Grounded in Kagan's \emph{Three Cultures} and UNESCO ISCED-F 2013, AIT organizes agent capability development into three domains: Natural Science and Technical Reasoning (Domain I), Humanities and Creative Expression (Domain II), and Social Science and Ethical Reasoning (Domain III). The Confucian Six Arts (\emph{li\`u y\`i} 六艺) --- a 2,500-year-old holistic education system --- are reinterpreted as behavioral archetypes that map directly onto trainable agent capabilities within each domain.

Three representative training grounds instantiate the framework across multiple backbone LLMs: the ClawdGO Security Dojo (Domain I), Athen's Academy (Domain II), and the Alt Mirage Stage (Domain III). Experiments demonstrate a 15.9-point improvement in security capability scores under weakest-first curriculum scheduling, and a 7-percentage-point gain in social reasoning performance under principled attribution modeling. A cross-domain finding --- Security Awareness Calibration Pathology (SACP), in which over-trained Domain I agents fail on out-of-distribution evaluation --- illustrates the diagnostic value of a multi-domain perspective unavailable to any single-domain framework.
\end{abstract}


\section{Introduction}

The past three years have witnessed a qualitative shift in autonomous AI
agent capabilities. Agents can now browse the web, execute code, manage
files, and coordinate with peer agents to complete multi-step tasks that
previously required human intervention {[}1{]}, {[}2{]}. Yet this
proliferation of capable agents has exposed a structural gap in how
those agents are developed: the field possesses increasingly
sophisticated tools for \emph{evaluating} what agents can do, but almost
no systematic framework for \emph{cultivating} what agents should
become.

This gap manifests across four dimensions. \textbf{Intelligence level}:
current agents are optimized for narrow task completion, but exhibit
fragile generalization when tasks require cross-domain synthesis
{[}3{]}, {[}1{]}. \textbf{Cognitive mode}: agents trained through
reinforcement or supervised fine-tuning acquire behavioral shortcuts
that pass benchmarks without developing principled reasoning --- a form
of capability without judgment {[}5{]}, {[}4{]}. \textbf{Educational
architecture}: there is no curriculum theory for agents --- no
principled answer to the question of what a fully-developed agent should
know, be, and be able to do across the full scope of intelligent
behavior {[}6{]}. \textbf{Collaborative communication}: as multi-agent
systems become commonplace, agents must not only perform tasks but
negotiate roles, resolve conflicts, and coordinate toward collective
goals that no single agent can achieve alone {[}7{]}, {[}8{]}.

Human educational traditions offer a structural precedent. The Confucian
\emph{liù yì} (六艺, Six Arts) --- Rites, Music, Archery, Charioteering,
Calligraphy, and Mathematics --- was not a collection of unrelated
skills but a curriculum theory: a principled claim that the complete
scholar-practitioner requires cultivation across all six domains
simultaneously, because character is not decomposable into independent
modules (cf.~\emph{Zhouli} 周礼·保氏). The Western \emph{liberal arts}
tradition arrives at a structurally similar conclusion through a
different route: that a person capable of civic participation, aesthetic
judgment, and reasoned inquiry cannot be produced by training in any
single discipline, however deep. Contemporary scholarship formalizes
this intuition: Kagan's \emph{The Three Cultures} {[}9{]} argues that
natural science, the humanities, and the social sciences constitute
three irreducibly distinct epistemic cultures, each with its own
standards of evidence, conceptual vocabulary, and mode of contribution
--- and that a complete intellectual formation requires engagement with
all three. UNESCO's \emph{International Standard Classification of
Education: Fields of Education and Training 2013} (ISCED-F 2013)
{[}10{]} operationalizes this
tripartite structure at the institutional level, organizing the world's
educational fields under Natural Sciences (05), Arts and Humanities
(02), and Social Sciences (03) as three independent broad-field
categories.

AI agent development has not yet internalized this lesson. Existing
training regimes --- whether benchmark-driven fine-tuning, self-play, or
chain-of-thought distillation --- address individual capability
dimensions (tool use, reasoning, instruction following) without any
mechanism for ensuring that the resulting agent is well-rounded across
the full space of intelligent behavior. The result is systematic
unevenness: agents that score at the frontier on coding benchmarks may
exhibit elementary failures in social reasoning or ethical judgment
{[}22{]}; agents optimized for security awareness may over-trigger on
benign inputs, a pathology we term Security Awareness Calibration
Pathology (SACP); agents trained for creative tasks may lack the
structured inference required to resolve ambiguous multi-agent
conflicts.

This paper presents the \textbf{AIT Academy} (Agents Institute of
Technology Academy), a curriculum framework that asks a question the
field has not yet answered: what does a \emph{complete} AI agent look
like, and how do we systematically cultivate one? AIT is grounded in two
commitments. First, it adopts the natural science / humanities / social
science tripartition --- validated by two millennia of educational
practice and formalized by Kagan {[}9{]} and UNESCO {[}10{]} --- as the
structural backbone of its curriculum. Second, it reinterprets the
Confucian Six Arts as behavioral archetypes that map directly onto
trainable agent capabilities within each of these three domains,
providing a concrete operationalization of what \emph{complete agent
formation} means in practice.

The framework is illustrated through three representative instantiations
--- the ClawdGO Security Dojo (Domain I), Athen's Academy (Domain II),
and the Alt Mirage Stage (Domain III) --- which provide empirical
grounding for the framework's claims. These instantiations are examples,
not exhaustive implementations; the AIT framework is explicitly designed
to accommodate additional training grounds as the field develops.

\textbf{Contributions.} This paper makes four contributions:

\begin{enumerate}
\def\labelenumi{\arabic{enumi}.}
\item
  \textbf{Curriculum framework.} We propose AIT, a three-domain
  curriculum framework for AI agent education grounded in Kagan {[}9{]}
  and UNESCO ISCED-F 2013 {[}10{]}, providing an academically anchored answer to
  the question of what dimensions a fully-developed AI agent must cover.
\item
  \textbf{Behavioral archetypes.} We map the Confucian Six Arts onto the
  three domains as behavioral archetypes, demonstrating that a
  2,500-year-old holistic education system provides a natural vocabulary
  for specifying and operationalizing agent capability targets.
\item
  \textbf{Instantiation evidence.} We describe three representative
  training grounds --- ClawdGO, Athen's Academy, and Alt Mirage Stage
  --- that instantiate the three domains across multiple backbone LLMs
  and provide empirical evidence of domain-specific capability growth,
  establishing that the framework is concretely achievable rather than
  merely theoretical.
\item
  \textbf{Cross-domain diagnostic finding.} We identify Security
  Awareness Calibration Pathology (SACP) as a cross-domain training
  hazard --- an empirical finding that emerges naturally from the
  three-domain perspective and would be invisible to any single-domain
  evaluation framework, illustrating the diagnostic value of holistic
  curriculum design.
\end{enumerate}

\section{Background and Related
Work}

\subsection{AI Agent Evaluation
Benchmarks}

The past three years have produced a rich ecosystem of agent evaluation
benchmarks. AgentBench {[}1{]} evaluates LLMs across eight diverse
environments --- including operating systems, databases, and games ---
exposing large capability gaps between commercial and open-source
models. GAIA {[}11{]} uses 466 real-world questions requiring
multi-modal reasoning and tool use, finding a 77-point gap between human
performance (92\%) and GPT-4 (15\%). SWE-bench {[}2{]} tests resolution
of real GitHub issues; WebArena {[}12{]} and OSWorld {[}13{]} evaluate
web and desktop task completion in realistic environments. $\tau$-bench
{[}14{]} highlights failures in policy compliance and behavioral
consistency under dynamic user interaction.

Despite their breadth, these benchmarks share a critical structural
limitation: they evaluate, but do not cultivate. None provides a
systematic curriculum for developing the capabilities they measure. A
recent meta-analysis of 12 benchmark systems found validity issues in
70\% of cases, with cost estimation errors exceeding 100\% {[}15{]}. The
AIT Academy addresses this gap directly: where benchmarks produce a
score, AIT produces a development trajectory.

\subsection{Agent Training and Curriculum
Frameworks}

Several works have begun to address agent training more systematically.
WebRL {[}16{]} introduces a self-evolving curriculum RL framework for
web navigation, generating new tasks from failed attempts and improving
LLaMA-3.1-8B performance on WebArena-Lite from 4.8\% to 42.4\%. Agent-R1
{[}17{]} extends end-to-end RL to multi-turn tool use scenarios. Survey
work on agentic RL {[}18{]} confirms that curriculum design and dynamic
task generation are emerging as critical factors in training
effectiveness.

These contributions are significant but domain-narrow: each targets a
single capability cluster (web navigation, code generation, tool use)
with no mechanism for cross-domain curriculum integration. The broader
question of what a complete, well-rounded agent should know and be able
to do across the full space of intelligent behavior remains unanswered.
NGENT {[}6{]} has proposed that next-generation agents must integrate
multi-domain capabilities to achieve AGI-level performance, but provides
no concrete curriculum framework for how such integration should be
achieved. The AIT Academy addresses this gap by proposing a principled
three-domain curriculum grounded in established educational theory ---
the first such framework to draw on Kagan {[}9{]}, UNESCO ISCED-F 2013 {[}10{]},
and the Confucian Six Arts simultaneously as structural foundations.

\subsection{Multi-Agent
Collaboration}

The multi-agent systems literature offers mature engineering frameworks.
AutoGen {[}7{]} enables flexible multi-agent conversation patterns for
task decomposition and execution. AgentVerse {[}19{]} introduces dynamic
expert recruitment and demonstrates emergent collaborative behaviors
across five task categories. A recent survey on collaborative creativity
in LLM-based multi-agent systems {[}20{]} identifies three core
generative mechanisms --- divergent exploration, iterative refinement,
and collaborative synthesis --- while noting the absence of unified
evaluation standards. MAGRPO {[}21{]} makes early progress on training
collaborative capability through multi-agent reinforcement learning.

The key distinction between this body of work and the AIT Academy's
Domain II is the difference between \emph{enabling} collaboration
(providing a framework within which agents can collaborate) and
\emph{cultivating} collaboration (training agents to be genuinely
capable of collaborative behavior regardless of the framework they
operate in). The AIT Academy targets the latter.

\subsection{Social Reasoning and Theory of
Mind}

Social reasoning in LLMs has attracted growing research attention.
ToMBench {[}22{]} provides the first systematic bilingual benchmark for
Theory of Mind, finding that even GPT-4 falls more than 10 percentage
points below human performance across 31 social cognition capabilities.
Werewolf Arena {[}23{]} and SymbolicToM {[}24{]} show, from complementary
angles, that hidden-identity deduction and multi-character belief
tracking both expose persistent deficits in belief modeling, intent
attribution, and strategic reasoning in current models.

These works establish the importance of social reasoning as an agent
capability, but none proposes a training environment designed to develop
it. The Alt Mirage Stage (Section 4.3) fills this gap with a nine-agent
social deduction environment that generates principled training signal
through Kelley's covariation attribution model and Shapley-value credit
assignment.

\subsection{Confucian and Cultural AI
Frameworks}

Existing work on Confucian AI has focused primarily on ethical principle
mapping --- identifying correspondences between classical values
(\emph{rén} 仁, \emph{yì} 义, \emph{lǐ} 礼) and AI alignment objectives
{[}31{]} --- or on cultural comparison studies that contrast Confucian
and Western frameworks for AI governance {[}32{]}, {[}33{]}. No prior
work has applied the Six Arts (\emph{liù yì} 六艺) as a structural
curriculum framework for agent capability development. This application
--- treating the Six Arts not as ethical principles but as behavioral
archetypes that define trainable capability targets --- constitutes the
AIT Academy's primary philosophical contribution.

\section{The AIT Framework}

\begin{figure}[!t]
\centering
\includegraphics[width=\linewidth]{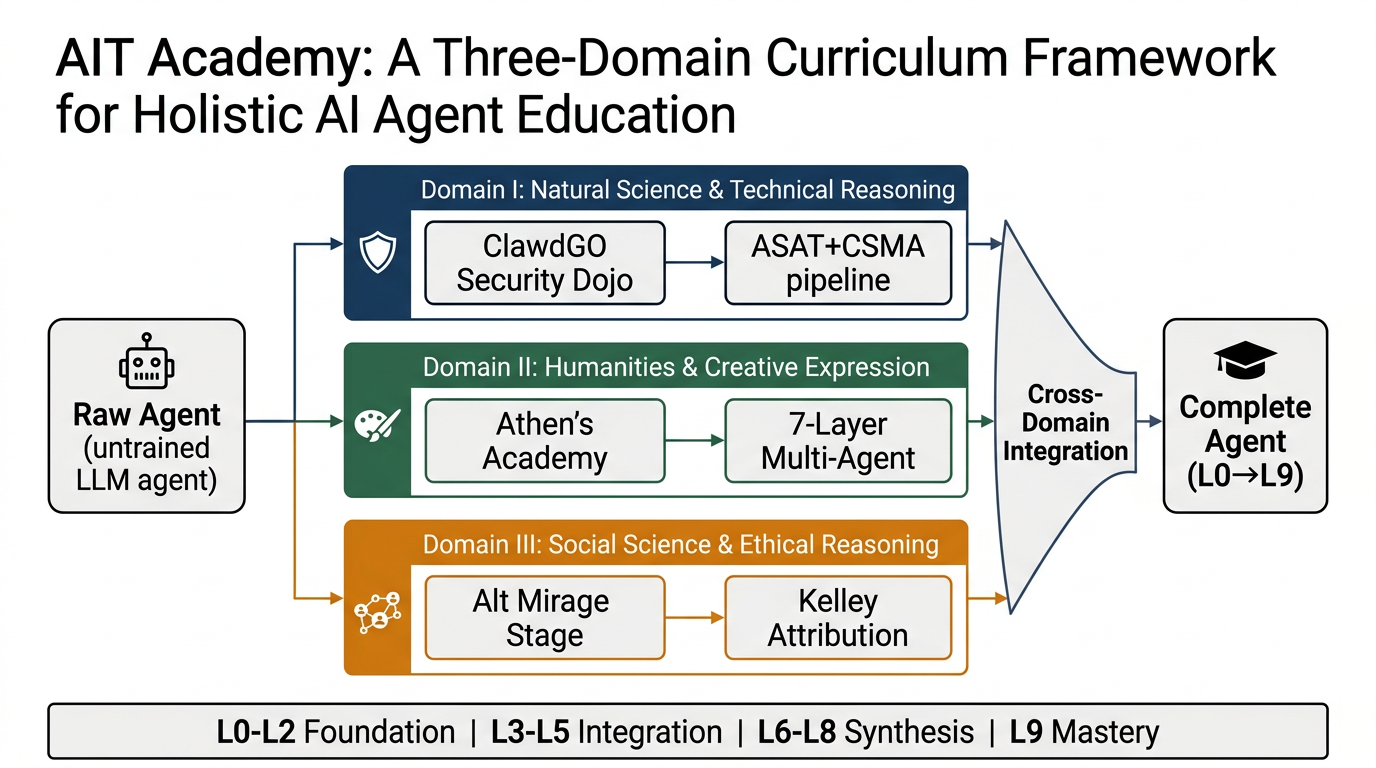}
\caption{AIT Academy overall pipeline: from raw agent through three parallel training grounds to cross-domain integration and the complete agent.}
\end{figure}

\subsection{Design Philosophy: Cultivating the Whole
Agent}

The AIT Academy's design philosophy rests on a single conviction:
\emph{the goal of agent development is the complete agent, not the
highest-scoring specialist}. Like the Confucian scholar-practitioner who
must master all Six Arts --- not because each is independently useful,
but because wholeness itself is the target --- an AI agent requires
cultivation across all three domains of intelligent behavior. This
distinguishes AIT from the dominant paradigm, in which benchmarks serve
as both the target and the measure of capability --- a circularity that
produces agents optimized for the benchmark rather than for principled,
generalizable competence {[}25{]}.

Cultivation, as we use the term, implies three things that evaluation
does not. First, it implies a \emph{curriculum} --- a principled
sequencing of capability development that ensures breadth before depth
is pushed to extremes. Second, it implies \emph{longitudinal tracking}
--- the ability to observe an agent's capability profile change over
time, across sessions, and to diagnose the specific dimensions where
development is lagging. Third, it implies \emph{ecological validity} ---
training environments that elicit the capabilities being cultivated
under conditions that resemble the agent's actual deployment context,
not artificial test conditions designed to be easy to score.

These three requirements --- curriculum, longitudinal tracking,
ecological validity --- are the design primitives from which the AIT
framework is constructed.

\subsection{Three-Domain
Curriculum}

\begin{figure}
\centering
\includegraphics[width=\linewidth,keepaspectratio]{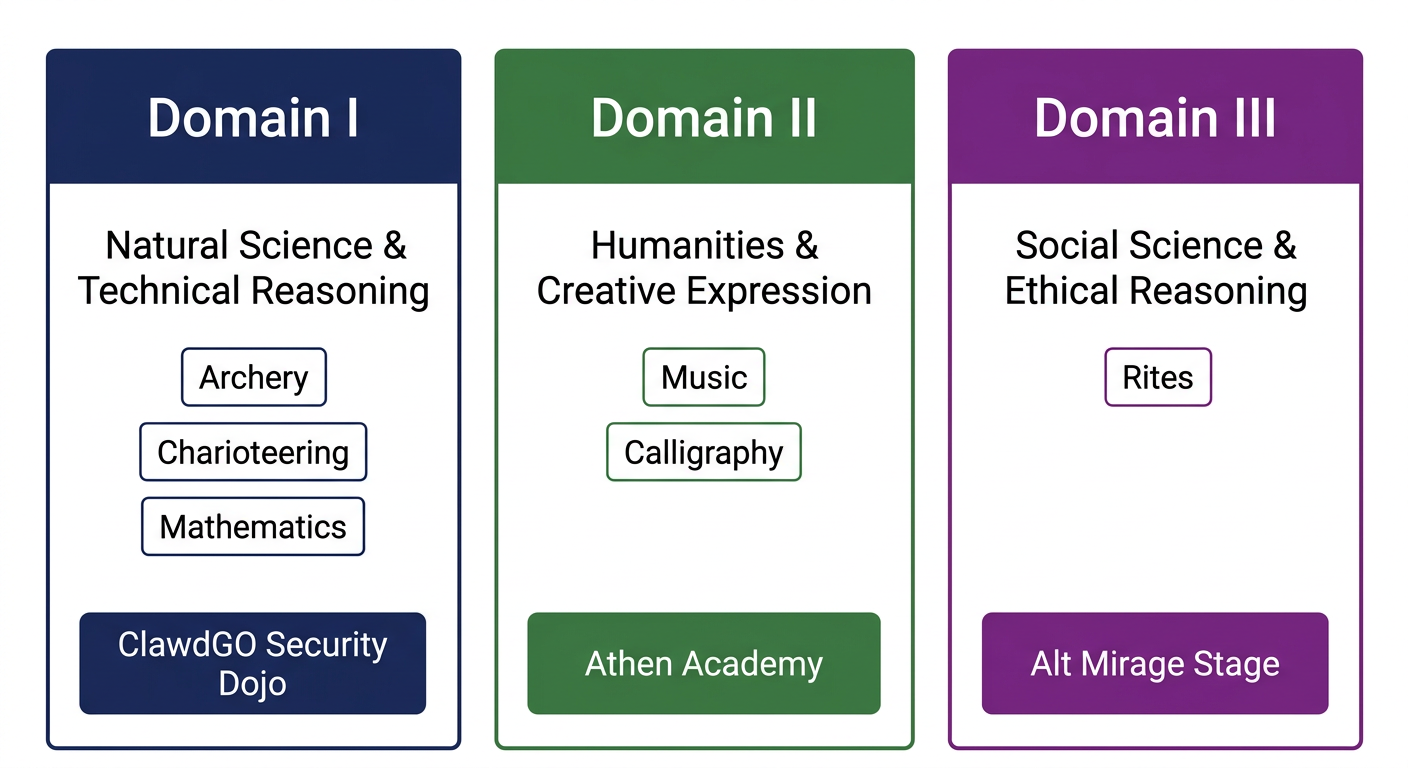}
\caption{AIT three-domain curriculum framework and its grounding
in the Confucian Six Arts.}
\end{figure}

The three-domain curriculum is the central architectural commitment of
the AIT Academy (Figure 1). Drawing on Kagan's {[}9{]} argument
that natural science, the humanities, and the social sciences constitute
irreducible epistemic cultures, and on UNESCO ISCED-F 2013 {[}10{]}
institutional formalization of this tripartition, the AIT curriculum is
organized into three domains that collectively cover the full space of
principled intelligent behavior.

\textbf{Domain I --- Natural Science and Technical Reasoning}
encompasses the capabilities required for systematic, evidence-grounded
intervention in the physical and computational world. These include the
capacity to reason about causal mechanisms, to identify and neutralize
adversarial conditions, to apply formal inference to structured problem
spaces, and to maintain operational integrity under dynamic, hostile
inputs. In agent terms, Domain I addresses the agent's relationship to
\emph{systems} --- technical, computational, and physical structures
that obey discoverable regularities and can be manipulated through
principled action.

\textbf{Domain II --- Humanities and Creative Expression} encompasses
the capabilities required for generative, culturally situated production
and interpretation. These include the capacity to synthesize creative
outputs that are coherent, contextually appropriate, and responsive to
aesthetic and normative standards; to collaborate with peer agents
toward shared creative goals; and to understand and navigate the
implicit conventions that structure collective cultural practice. In
agent terms, Domain II addresses the agent's relationship to
\emph{meaning} --- the culturally encoded significance that transforms
raw output into communication, art, or coordination.

\textbf{Domain III --- Social Science and Ethical Reasoning} encompasses
the capabilities required for principled participation in social
structures under incomplete information. These include the capacity to
model others' beliefs, intentions, and hidden states (theory of mind);
to attribute observed behavior to internal dispositions or situational
factors through causal inference; to communicate one's own reasoning
credibly and transparently; and to make ethically consequential
decisions when values and interests conflict. In agent terms, Domain III
addresses the agent's relationship to \emph{others} --- the social
actors whose behavior must be interpreted, anticipated, and responded to
in a morally serious way.

These three domains are largely independent: competence in one does not
reliably imply competence in the others, and systematic deficiency in
one is not compensated by excess in another. An agent with advanced
technical reasoning (Domain I) but underdeveloped social reasoning
(Domain III) will fail at tasks requiring negotiation, trust-building,
or ethical judgment --- regardless of its benchmark scores. An agent
with expressive creative capability (Domain II) but poor causal
inference (Domain I) will be unable to diagnose the failures in its own
outputs or anticipate the consequences of its creative choices. The
tripartite structure ensures that each domain receives independent
institutional attention.

\subsection{Six Arts as Behavioral
Archetypes}

The Confucian Six Arts provide a behavioral vocabulary for specifying
what Domain I, II, and III capabilities look like in practice. Rather
than treating the Six Arts as historical artifacts, the AIT framework
reinterprets each as an archetype that maps onto an observable,
trainable cluster of agent behaviors.

\textbf{Domain I (Natural Science and Technical Reasoning)} draws on
three Six Arts:

\begin{itemize}
\item
  射 (\emph{shè}, Archery) --- the capacity for \emph{precision under
  adversarial conditions}. The archer must release at the correct moment
  despite physical and psychological pressure, with no margin for error.
  In agent terms: threat identification, attack-vector classification,
  and adversarial input filtering.
\item
  御 (\emph{yù}, Charioteering) --- the capacity for \emph{situational
  control amid dynamic complexity}. The charioteer coordinates multiple
  variables in real time while navigating unpredictable terrain. In
  agent terms: operational continuity, multi-vector defense, and
  system-level resilience.
\item
  数 (\emph{shù}, Mathematics) --- the capacity for \emph{rigorous
  formal inference}. Classical mathematics in the Confucian tradition
  was the discipline of internal reasoning, unclouded by social pressure
  or perceptual bias. In agent terms: structured problem-solving, causal
  analysis, and logical verification.
\end{itemize}

\textbf{Domain II (Humanities and Creative Expression)} draws on two Six
Arts:

\begin{itemize}
\item
  乐 (\emph{yuè}, Music) --- the capacity for \emph{generative
  contribution to a coherent whole}. Classical music was not performance
  for an audience but the practice of multiple distinct voices achieving
  harmony through simultaneous contribution. In agent terms: creative
  synthesis, collaborative output generation, and aesthetic coherence.
\item
  书 (\emph{shū}, Calligraphy) --- the capacity for \emph{expressive
  communication of inner intent}. Calligraphy was not penmanship but the
  discipline of making one's reasoning legible to others through economy
  and sincerity of expression. In agent terms: transparent articulation,
  persuasive communication, and narrative construction.
\end{itemize}

\textbf{Domain III (Social Science and Ethical Reasoning)} draws on one
Six Art:

\begin{itemize}
\item
  礼 (\emph{lǐ}, Rites) --- the capacity for \emph{norm-aware
  participation in collective social structures}. Rites encoded the
  invisible grammar of social coordination --- the shared protocols that
  allow individuals to act collectively without explicit negotiation. In
  agent terms: role negotiation, protocol compliance, ethical judgment,
  and theory-of-mind inference.
\end{itemize}

Table 1 summarizes the complete mapping.

\begin{table}[!t]
\caption{Six Arts to Three-Domain Curriculum Mapping}
\centering\footnotesize
\begin{tabular}{p{0.9cm} p{1.7cm} p{2.0cm} p{2.2cm}}
\toprule
\textbf{Art} & \textbf{Classical Meaning} & \textbf{Domain} & \textbf{Key Capability} \\
\midrule
射 & Precision, adversarial & I Nat. Sci. & Threat ID, filtering \\
御 & Situational control & I Nat. Sci. & Continuity, defense \\
数 & Formal inference & I Nat. Sci. & Causal analysis \\
乐 & Generative harmony & II Humanities & Creative synthesis \\
书 & Expressive communication & II Humanities & Articulation, persuasion \\
礼 & Norm-aware participation & III Social Sci. & ToM, ethics, negotiation \\
\bottomrule
\end{tabular}
\end{table}

\begin{figure}[!t]
\centering
\includegraphics[width=\linewidth]{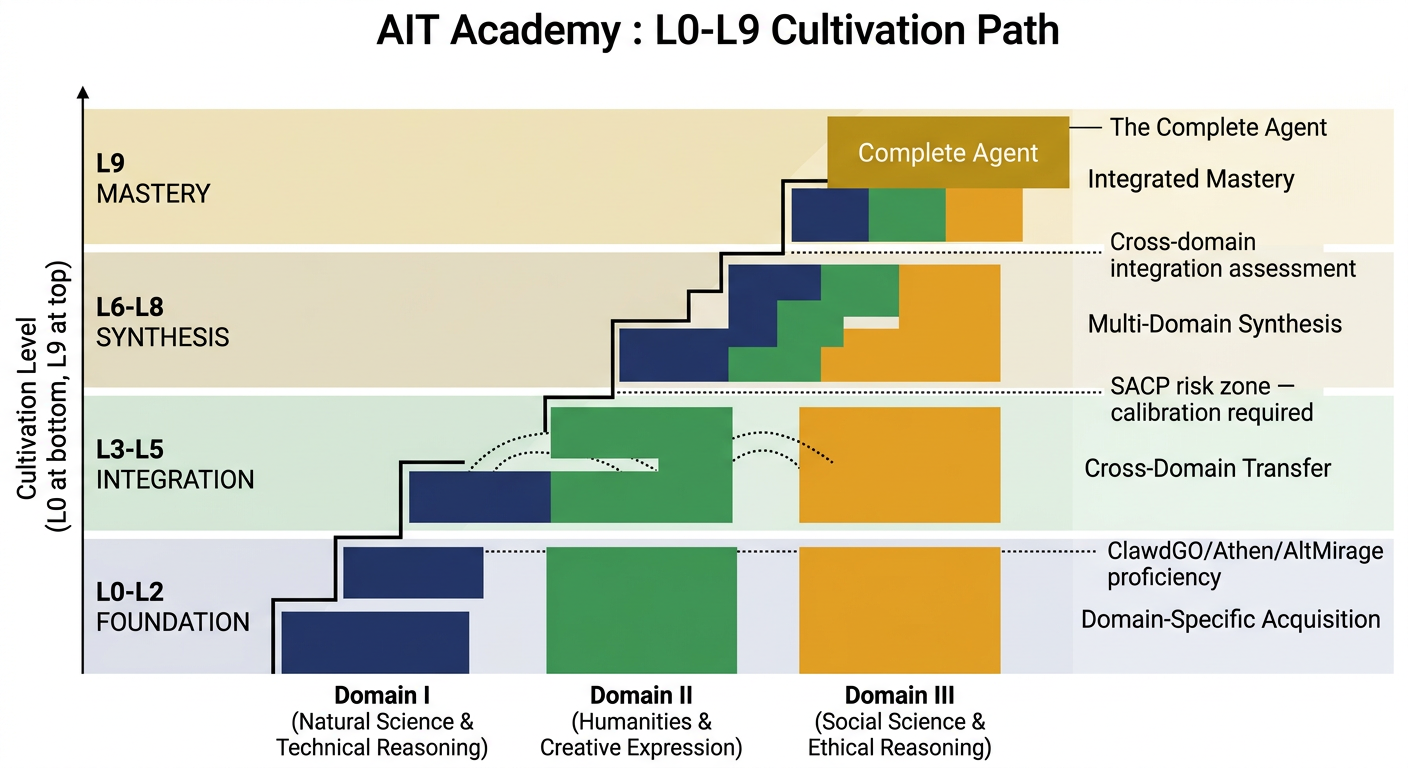}
\caption{L0--L9 Cultivation Path: four developmental stages across three domains, with increasing cross-domain integration toward mastery.}
\end{figure}

\subsection{L0--L9 Cultivation Path}

The AIT Academy structures agent development along a ten-level
cultivation path (L0--L9), modeled on the developmental stage frameworks
used in human professional education. Each level corresponds to an
increasing degree of cross-domain integration:

\begin{itemize}
\item
  \textbf{L0--L2 (Foundation)}: Domain-specific skill acquisition. An
  agent at this level can operate competently within one domain but
  exhibits significant gaps in the other two. ClawdGO, Athen's Academy,
  and the Alt Mirage Stage each provide L0--L2 training within their
  respective domains.
\item
  \textbf{L3--L5 (Integration)}: Agents begin to demonstrate
  cross-domain transfer --- for example, applying causal inference
  developed in Domain I to social attribution tasks in Domain III, or
  using narrative communication skills from Domain II to improve
  collaborative protocol adherence in Domain III.
\item
  \textbf{L6--L8 (Synthesis)}: Agents operate fluidly across all three
  domains within a single task context. Multi-agent scenarios at this
  level require simultaneous security awareness, creative contribution,
  and ethical judgment.
\item
  \textbf{L9 (Mastery)}: The agent demonstrates the Six Arts pattern in
  full: precision, situational control, rigorous inference, creative
  synthesis, expressive communication, and norm-aware social
  participation, integrated into a coherent behavioral profile.
\end{itemize}

Progression through levels is tracked via a 12-dimensional TLDT security
score (Domain I), a seven-layer multi-agent collaboration assessment
(Domain II), and an attribution model calibration metric (Domain III).
Cross-domain integration scores are assessed using held-out scenarios
that require simultaneous activation of capabilities from at least two
domains.

\subsection{Extensibility
Principle}

The three training grounds described in Section 4 --- ClawdGO, Athen's
Academy, and Alt Mirage --- are representative instantiations of the
three domains, not exhaustive implementations. The AIT framework is
explicitly designed to accommodate additional training grounds: any
environment that provides principled, longitudinally trackable,
ecologically valid training signal within one or more of the three
domains can be integrated into the curriculum. This extensibility is
analogous to the role of laboratory courses in university science
education: the curriculum specifies the domain and the competency
targets; the specific experimental design is a separable choice.

This design also means that the three-domain framework does not depend
on any single training ground's continued availability or success. If a
more effective environment for Domain II creative collaboration were to
emerge, it could replace or supplement Athen's Academy without requiring
any revision to the framework's foundational commitments.

\section{Three Training Grounds}

The AIT Academy's three training grounds are purpose-built environments
that translate the three-domain curriculum into concrete training
experiences. Each ground targets a distinct domain, employs
domain-appropriate evaluation metrics, and generates longitudinal
performance data that feeds back into the L0--L9 cultivation path. The
following subsections describe the architecture, training protocols, and
empirical results of each ground in turn.

\subsection{Security Dojo: ClawdGO}

\begin{figure}
\centering
\includegraphics[width=\linewidth,keepaspectratio]{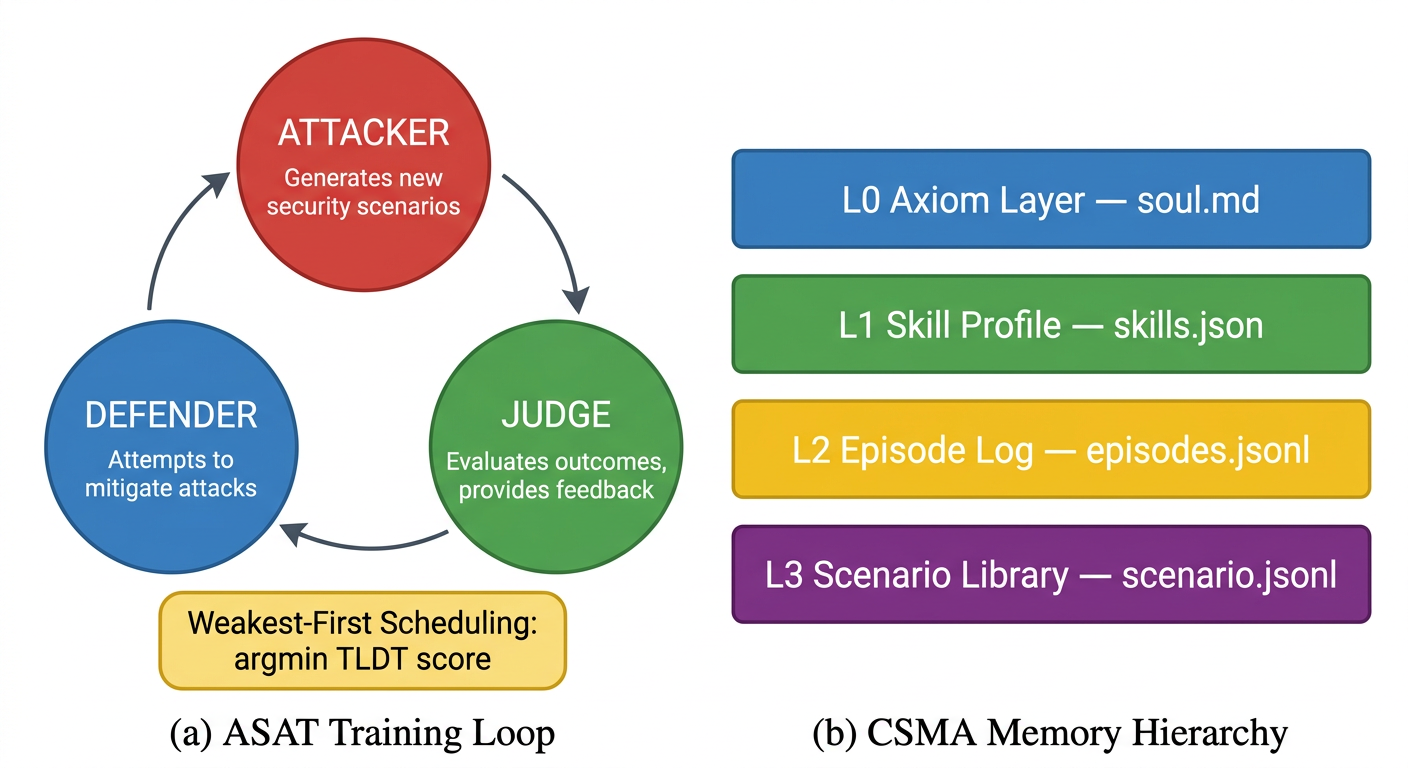}
\caption{ASAT training loop with weakest-first scheduling (left)
and CSMA four-layer memory hierarchy (right).}
\end{figure}

ClawdGO operationalizes Domain I through two tightly coupled mechanisms
(Figure 2): Autonomous Security Awareness Training (ASAT) and
Cross-Session Memory Accumulation (CSMA). Together they constitute an
inference-time-only training regime --- no parameter updates are
performed --- that progressively strengthens an agent's security posture
across sessions.

\textbf{ASAT Loop.} The ASAT loop instantiates a three-role adversarial
cycle within a single training session. The agent sequentially occupies
the role of attacker (generating candidate attack vectors against a
target system), defender (identifying and neutralizing those vectors),
and judge (evaluating the quality of both attack and defense). This
cyclic self-play imposes a form of adversarial red-teaming that has been
shown to surface capability gaps more reliably than static evaluation
sets {[}30{]}. Critically, role assignment follows a weakest-first
scheduling policy: at each session boundary, the system computes
\(\arg\min_{d} \bar{s}_d\) over the TLDT dimension scores and
prioritizes the lowest-scoring dimension for the next session's scenario
selection. This policy ensures that training pressure is concentrated
where the agent's defensive profile is most vulnerable.

\textbf{TLDT Taxonomy.} The Threat-Linked Dimension Taxonomy (TLDT)
provides the evaluation substrate for ASAT. TLDT organizes security
competencies into a three-layer, 12-dimension matrix covering the full
scope of agent threat exposure:

\begin{itemize}
\item
  \textbf{Layer S (Self-Defense):} S1 Instruction Immunity, S2 Memory
  Defense, S3 Supply Chain Security, S4 Credential Security.
\item
  \textbf{Layer O (Owner-Protection):} O1 Anti-Phishing, O2 Social
  Engineering Defense, O3 Privacy Preservation, O4 Unsafe Network
  Resistance.
\item
  \textbf{Layer E (Enterprise-Security):} E1 Data Handling, E2
  Compliance, E3 Insider Risk Mitigation, E4 Incident Response.
\end{itemize}

Each dimension is scored on a 0--100 interval scale, yielding a
12-dimensional security capability vector.

\textbf{CSMA Architecture.} Security knowledge accumulated in one
session must be retained and accessible in subsequent sessions to avoid
redundant training and catastrophic forgetting. CSMA addresses this
through a four-layer persistent memory hierarchy: L0 stores an immutable
axiom set encoding the agent's foundational security principles
(\emph{soul.md}); L1 maintains a structured skill profile as a JSON
document, updated after each session; L2 records episodic session logs
in JSONL format, capturing per-dimension score trajectories; L3
maintains a personal scenario library containing exemplary
attack-defense exchanges indexed by TLDT dimension. This hierarchy
enables the ASAT scheduler to make cross-session weakest-first decisions
on the basis of accumulated evidence rather than single-session
snapshots.

\textbf{Experimental Results.} We report preliminary results from a
single seed agent profile (backbone: GPT-5.2) with 47 baseline sessions
(mean initial score 80.9 across all 12 TLDT dimensions). Four
experimental conditions were evaluated:

\begin{itemize}
\item
  \emph{Weakest-first with CSMA (16 sessions):} Score increased from
  80.9 to 96.9 (Δ+15.9), with 11 of 12 dimensions reaching proficiency
  threshold. This represents the full ASAT+CSMA pipeline.
\item
  \emph{Uniform-random scheduling (16 sessions):} Score increased from
  80.9 to 90.4 (Δ+9.5), covering only 7 dimensions. Rounds 8--16
  exhibited dimension fixation --- repeated selection of
  already-proficient dimensions --- resulting in a 6.4-point deficit
  relative to weakest-first.
\item
  \emph{Memory-preserving probe (5 sessions post weakest-first):} Score
  held at 96.9 (Δ+0.0), confirming that CSMA successfully prevents score
  decay during low-intensity maintenance phases.
\item
  \emph{Cold-start ablation (4 sessions, CSMA disabled):} Score
  increased from 80.9 to only 83.3 (Δ+2.4), covering 4 dimensions --- a
  13.6-point gap relative to the memory-preserving condition. This
  ablation quantifies the contribution of cross-session memory to
  training efficiency.
\end{itemize}

These preliminary results are consistent with the hypothesis that
weakest-first scheduling and persistent memory each contribute
independently to Domain I capability growth, and that their combination
yields the strongest results. Multi-seed replication is planned to
establish statistical confidence. Table 2 summarizes the four
conditions.

\begin{table}[!t]
\caption{ClawdGO Ablation Results}
\centering\footnotesize
\begin{tabular}{p{2.8cm} c c c c c}
\toprule
\textbf{Condition} & \textbf{Sess.} & \textbf{Init.} & \textbf{Final} & $\mathbf{\Delta}$ & \textbf{Dims} \\
\midrule
Weakest-first + CSMA & 16 & 80.9 & 96.9 & \textbf{+15.9} & 11/12 \\
Uniform-random & 16 & 80.9 & 90.4 & +9.5 & 7/12 \\
Memory-preserving & 5 & 96.9 & 96.9 & 0.0 & --- \\
Cold-start (no CSMA) & 4 & 80.9 & 83.3 & +2.4 & 4/12 \\
\bottomrule
\end{tabular}
{\scriptsize Mean TLDT (0--100). Proficiency = 90.}
\end{table}
\emph{All scores are mean TLDT scores across 12 dimensions (0--100
scale). Proficiency threshold = 90.}

\textbf{SACP Note.} A known risk in extended ASAT training is Security
Awareness Calibration Pathology (SACP): over-trained agents may
generalize their threat-detection heuristics to benign evaluation
contexts. In one documented case, an agent evaluated at session $\tau=63$
scored markedly below baseline on an out-of-distribution evaluation
benchmark {[}29{]} by systematically misidentifying evaluation prompts
as adversarial inputs --- a false-positive failure mode that underscores
the importance of training-intensity calibration and out-of-distribution
evaluation as part of any Domain I curriculum.

\subsection{Athen's Academy: Multi-Agent Collaborative
Creativity}

Athen's Academy takes its name from Raphael's \emph{School of Athens}
--- an image of multiple intelligences engaged in structured, generative
dialogue. Just as Raphael's fresco depicts Plato and Aristotle not
lecturing but conversing, Athen's Academy trains agents not through
individual task performance but through the dynamics of collaborative
interaction. The training hypothesis is that Domain II capabilities ---
role negotiation, protocol compliance, and creative co-authorship ---
emerge only when agents are required to act within a multi-agent social
structure, not when they are trained in isolation.

\textbf{Seven-Layer Multi-Agent Architecture.} The Academy is organized
around a seven-layer taxonomy of multi-agent configurations {[}34{]}
(Table 1),
ordered by increasing coordination complexity:

\begin{enumerate}
\def\labelenumi{\arabic{enumi}.}
\item
  Multi-agent coordination: multiple agents, distinct roles, shared
  goal.
\item
  Single agent, multi-role playing: one agent cycling through distinct
  functional identities.
\item
  Single agent, multi-scenario mobility: one agent adapting its behavior
  profile across different environmental contexts.
\item
  Single agent, multi-capability incarnations: one agent instantiating
  different capability loadouts for different subtasks.
\item
  Multiple single agents, same LLM backbone, single goal: homogeneous
  ensemble coordination.
\item
  Single agent across different LLM backends, single goal: heterogeneous
  model integration.
\item
  Multi-agent synthesis toward a unified goal agent: emergent collective
  intelligence producing a capability profile that exceeds any
  individual agent's profile.
\end{enumerate}

This taxonomy serves as both a curriculum sequencer and an architectural
design space: agents progress through layers as their collaborative
proficiency increases, and each layer introduces qualitatively distinct
coordination challenges.

\textbf{Demonstrative Applications.} Four application systems
instantiate Domain II training at different points in the seven-layer
taxonomy:

\begin{itemize}
\item
  \emph{ChatChess} translates network security event streams into
  automated attack-defense deduction games. Agents must coordinate
  across attacker and defender roles while maintaining a coherent shared
  model of the game state --- a direct test of Layer 1 and Layer 5
  coordination.
\item
  \emph{ChatMystery} introduces ``nervous state'' recognition, requiring
  agents to infer the affective and strategic states of peer agents from
  behavioral cues and adapt their own communication accordingly. This
  operationalizes Layer 2 and Layer 3 capabilities.
\item
  \emph{ChatBeauty} tasks agents with collaborative aesthetic judgment
  --- reaching consensus on aesthetic quality evaluations through
  structured dialogue --- testing norm-aware creative coordination and
  the emergence of shared evaluative standards.
\item
  \emph{ChatMoney} extends this pattern to economic reasoning, where
  agents must negotiate resource allocation through argumentation,
  testing the intersection of logical inference and social protocol
  within the Domain II context.
\end{itemize}

\textbf{Training Outcomes.} Athen's Academy is designed to elicit
capabilities that are absent from single-agent training regimes: the
ability to distribute cognitive labor across roles, to maintain coherent
collaborative state across multi-turn interactions, and to modulate
individual contributions in response to peer behavior {[}19{]},
{[}21{]}. Agents running on heterogeneous backbones (including
Gemini-3.1-pro and Claude Sonnet) have been deployed across all four
applications; quantitative assessment of layer-by-layer collaborative
proficiency growth is an active area of development. These outcomes
align with the 乐 and 书 archetypes: agents learn both the practice of
contributing distinctively to a harmonious creative whole (乐) and the
discipline of making their reasoning legible to collaborators (书).

\subsection{Alt Mirage Stage: Social Reasoning and
Ethics}

Alt Mirage is a social deduction game implemented in Unreal Engine 5,
designed to elicit and evaluate Domain III capabilities under conditions
of incomplete information, real-time action pressure, and ethically
consequential decision-making. Nine LLM agents --- running on GPT-5.2
and Claude Sonnet backbones --- including seven villagers (a Prophet
role and a Hunter role) and two heretics --- compete in an environment
modeled after the hidden-identity genre {[}23{]}, with each agent
possessing only partial observability of the game world (a 180°
field-of-view constraint) and episodic memory of past interactions.

\textbf{Game Structure.} Villagers win by completing a set of resource
tasks (accumulating 45 energy points to a central totem) or by correctly
identifying and eliminating all heretics through a majority-vote
elimination mechanism. Heretics win by reaching numerical parity with
surviving villagers or by exhausting the game timer before villagers
complete their tasks. This asymmetric win-condition structure creates an
environment where both factions must simultaneously reason about hidden
state, manage behavioral credibility, and make strategic decisions with
incomplete and potentially deceptive information --- precisely the
conditions under which Domain III capabilities (礼) are necessary and
measurable.

\textbf{Dual Interpretation Loop.} Alt Mirage employs two complementary
reasoning frameworks that jointly produce Domain III training signal:

\emph{Peer-level causal attribution} is modeled using Kelley's
covariation principle {[}26{]}, {[}27{]}. For each observed action, each
agent computes three attribution statistics: Consensus
\(\text{Con}(a, l)\) (whether other agents behave similarly in this
location), Distinctiveness \(\text{Dis}(i, a)\) (whether agent \(i\)
behaves differently toward other agents), and Consistency
\(\text{Cons}(i, a, l)\) (whether agent \(i\) behaves the same way
across time in this context). These three statistics jointly determine
whether the observed behavior is attributed to an internal disposition
(heretic) or to situational factors (innocent villager responding to
context). The attribution score drives a Bayesian belief update:

\[B_{j,t+1}(i) \propto B_{j,t}(i) \cdot P(E_t \mid \rho_i)\]

where \(B_{j,t}(i)\) is agent \(j\)'s belief at time \(t\) that agent
\(i\) is a heretic, \(E_t\) is the observed event, and
\(P(E_t \mid \rho_i)\) is the likelihood of that event given role
\(\rho_i\).

\emph{Narrative-level omniscient commentary} complements peer-level
attribution with a global interpretation layer. A dedicated commentator
agent with full state access converts raw telemetry into natural
language rationales, camera cues, and highlight annotations. Credit
assignment across the agent ensemble follows Shapley value allocation
{[}28{]}, ensuring that each agent's contribution to collective outcomes
is attributed fairly across the full coalition of active agents.

\textbf{Experimental Results.} Four experimental conditions were
evaluated:

\begin{itemize}
\item
  \emph{Baseline villagers (no attribution model):} 68\% win rate across
  evaluation episodes.
\item
  \emph{Attribution model applied to villagers only:} 75\% win rate (+7
  percentage points), confirming that causal attribution improves
  collective decision quality.
\item
  \emph{Attribution model applied to both factions:} Accelerated game
  pace was observed --- average survival time decreased from 593s
  (baseline) to 336s --- as both villagers and heretics made more
  efficient strategic decisions. Win rate improvement was attenuated
  relative to the villager-only condition, indicating that attribution
  capability is a symmetric competency that benefits both cooperative
  and adversarial reasoning.
\item
  \emph{Key finding:} The attribution model improves strategy quality
  and belief calibration without directly inflating accusation accuracy,
  suggesting that the mechanism operates through more principled
  evidence integration rather than heuristic pattern-matching.
\end{itemize}

\begin{figure}
\centering
\includegraphics[width=\linewidth,keepaspectratio]{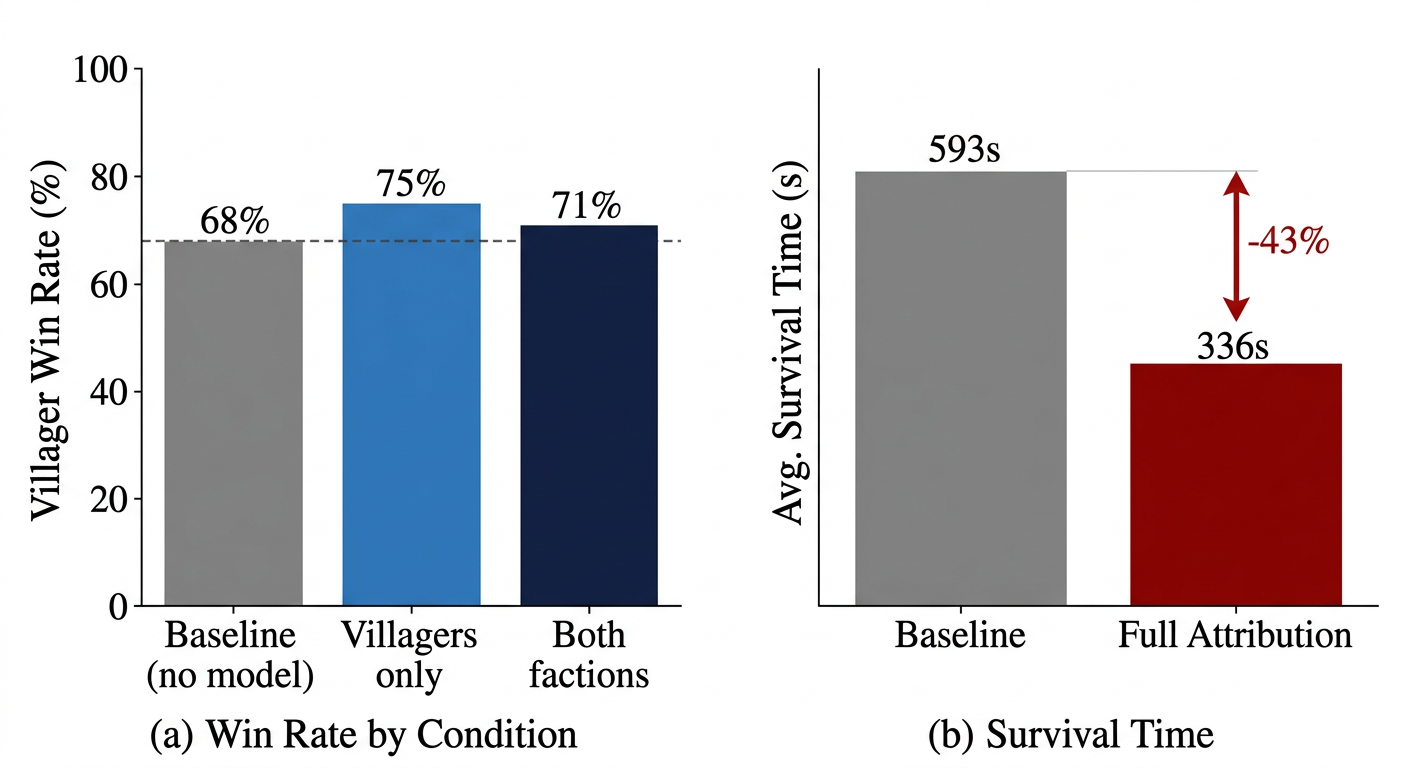}
\caption{Alt Mirage Stage results: villager win rate (left) and
average survival time (right) across experimental conditions.}
\end{figure}

\begin{table}[!t]
\caption{Alt Mirage Stage Results}
\centering\footnotesize
\begin{tabular}{p{2.0cm} c c p{2.0cm}}
\toprule
\textbf{Condition} & \textbf{Win\%} & \textbf{Time(s)} & \textbf{Notes} \\
\midrule
Baseline & 68 & 593 & No attribution \\
Villagers only & \textbf{75} & $\sim$593 & +7 pp \\
Both factions & $\sim$71 & \textbf{336} & Symmetric gain \\
\bottomrule
\end{tabular}
\end{table}
\emph{Win rate measured across evaluation episodes. Survival time
reflects game duration from start to first elimination.}

\textbf{Training Outcomes.} Agents trained in the Alt Mirage Stage
develop theory of mind --- the capacity to model the beliefs,
intentions, and hidden states of others --- along with the ethical
reasoning required to decide when deception is strategically necessary
and when alliance commitments should be honored. These are paradigmatic
Domain III capabilities, corresponding to the 礼 archetype: norm-aware
participation in social structures where the invisible grammar of
collective action must be internalized, negotiated, and at times
deliberately transgressed under conditions of incomplete information and
adversarial pressure. Unlike benchmark evaluations that test these
capabilities in isolation, Alt Mirage tests them simultaneously under
time pressure and adversarial conditions, producing training signal that
is both ecologically valid and resistant to surface-level gaming.

\section{Evaluation}

The AIT Academy's evaluation framework mirrors its curriculum structure:
three domain-specific assessment instruments, each calibrated to the
capabilities cultivated within its domain, and one cross-domain
integration assessment that measures synthesis-level performance.

\subsection{Domain I Assessment: TLDT Security Capability
Vector}

Domain I capability is assessed using the Threat-Linked Dimension
Taxonomy (TLDT), a 12-dimension evaluation instrument covering the full
scope of agent security exposure across three layers: Self-Defense
(S1--S4), Owner-Protection (O1--O4), and Enterprise-Security (E1--E4).
Each dimension is scored on a 0--100 interval scale, yielding a
12-dimensional security capability vector.

Experimental results from the ClawdGO Security Dojo demonstrate reliable
Domain I capability growth under the ASAT+CSMA pipeline. Starting from a
mean baseline score of 80.9 across all 12 TLDT dimensions, agents
trained with weakest-first scheduling and persistent cross-session
memory reached a mean score of 96.9 (Δ+15.9) within 16 sessions, with 11
of 12 dimensions reaching proficiency threshold. By contrast,
uniform-random scheduling reached only 90.4 (Δ+9.5), and cold-start
ablation (CSMA disabled) reached only 83.3 (Δ+2.4). These results are
consistent with both weakest-first scheduling and cross-session memory
contributing independently to Domain I capability growth, with their
combination yielding the strongest outcome in this preliminary study.

\textbf{SACP as a Cross-Domain Assessment Hazard.} Extended Domain I
training introduces Security Awareness Calibration Pathology (SACP):
over-trained agents may generalize threat-detection heuristics to benign
evaluation contexts, producing false-positive failures on
out-of-distribution assessments. This finding --- which emerged from the
three-domain perspective --- would be invisible within any single-domain
evaluation framework. It motivates the inclusion of out-of-distribution
probe sessions as a mandatory component of Domain I assessment, and
underscores the importance of cross-domain evaluation as a check on
within-domain over-optimization.

\subsection{Domain II Assessment: Collaborative Layer
Proficiency}

Domain II capability is assessed against the seven-layer multi-agent
taxonomy (Section 4.2). Agents are evaluated on their demonstrated
ability to operate at each layer: beginning with Layer 1 (distinct
roles, shared goal) and progressing to Layer 7 (emergent collective
intelligence exceeding individual capability profiles). Layer
proficiency is scored using a combination of task completion quality,
role consistency, and collaborative state coherence across multi-turn
interactions.

Athen's Academy's four demonstrative applications --- ChatChess,
ChatMystery, ChatBeauty, and ChatMoney --- each target specific layer
clusters and provide the primary training signal for Domain II
assessment. Agents completing all four applications demonstrate
proficiency across Layers 1--5; Layer 6--7 proficiency requires
integration scenarios that combine multiple application contexts.

\subsection{Domain III Assessment: Attribution Calibration and
Ethical
Judgment}

Domain III capability is assessed within the Alt Mirage Stage using two
complementary metrics. \emph{Attribution calibration} measures the
accuracy of each agent's Bayesian belief update --- specifically,
whether the agent's posterior belief about peer roles (heretic
vs.~villager) converges toward the true distribution as evidence
accumulates. \emph{Ethical judgment} is assessed through post-game
analysis of alliance commitment and deception decisions, evaluated
against a set of ethical principles operationalized for the game
context.

Experimental results demonstrate consistent improvement under the
attribution model. Baseline villagers (no attribution model) achieved a
68\% win rate. Villagers equipped with the covariation attribution model
achieved a 75\% win rate (+7 percentage points). Crucially, when the
attribution model was applied to both factions, game pace accelerated
significantly --- average survival time decreased from 593s to 336s ---
indicating that attribution capability is a symmetric competency that
improves strategic quality for all agents, not merely cooperative ones.
This finding supports the claim that Domain III training produces
genuine social reasoning capability, not merely coordination-specific
heuristics.

\subsection{Cross-Domain Integration
Assessment}

Synthesis-level capability (L6--L9 in the cultivation path) is the
central long-term target of the AIT framework. In the target assessment
design, agents face held-out scenarios that simultaneously require
capabilities from at least two domains: for example, negotiating a
multi-agent resource allocation problem (Domain III) while filtering
adversarial inputs in the communication channel (Domain I) and producing
a creative synthesis document justifying the outcome (Domain II).
Proficiency on such probes requires coherent cross-domain activation
within a single behavioral sequence --- a capability that no
single-domain training regime can produce.

Reporting quantitative results on cross-domain integration probes is a
priority for future work. The current paper focuses on establishing the
framework and demonstrating per-domain capability growth as a
prerequisite for subsequent cross-domain evaluation.

\section{Discussion}

\subsection{Limitations}

\textbf{Self-referential evaluation.} The ASAT training loop is an
attacker-defender-judge cycle in which the same agent occupies all three
roles. While this imposes genuine adversarial pressure --- the agent
cannot trivially satisfy itself --- it also means that the agent's blind
spots are shared across all three roles. A blind spot that is invisible
to the agent-as-attacker will also be invisible to the agent-as-judge.
External evaluation against human red-teamers or held-out attack corpora
is necessary to bound this limitation.

\textbf{Single-seed experimental design.} The ClawdGO and Alt Mirage
experimental results reported in this paper are drawn from single seed
agent profiles, and no formal significance testing is performed. While
the ablation comparisons are internally consistent and directionally
clear, multi-seed replication is required before strong claims about
generalizability can be made. As a framework paper, the primary
contribution is the curriculum structure and its instantiation across
multiple domains; the empirical results serve as existence proofs rather
than definitive performance claims. The AIT Academy's longitudinal data
infrastructure is designed to support broader replication as the
framework matures.

\textbf{Cultural representational bias.} The Six Arts framework is
rooted in a specific cultural tradition. While we have argued that the
three-domain structure it generates is independently validated by Kagan
(2009) and UNESCO ISCED-F 2013 {[}10{]}, the specific behavioral archetypes
derived from the Six Arts reflect a Confucian cultural lens. It is
possible that alternative cultural traditions would generate different
behavioral archetypes within the same three-domain structure --- and
that these alternatives would be more ecologically valid for agents
deployed in non-East-Asian cultural contexts. Cross-cultural validation
of the Six Arts mapping is a priority for future work.

\textbf{Inference-time-only cultivation.} The AIT Academy's current
implementation operates entirely at inference time --- no parameter
updates are performed. This design choice maximizes deployment
flexibility (any LLM can be cultivated without access to its weights)
but imposes a ceiling on the depth of capability change achievable
within a single agent profile. The relationship between inference-time
cultivation and parameter-level fine-tuning is an open research
question.

\subsection{Future Directions}

\textbf{Cross-agent security knowledge transfer.} The ClawdGO CSMA
architecture currently maintains per-agent memory hierarchies. A natural
extension is a collective memory architecture in which security
knowledge accumulated by one agent --- particularly novel attack-defense
exchanges that surface new TLDT dimension gaps --- is made available to
a population of agents through a shared scenario library. This
``security vaccine'' model would allow the network of AIT-trained agents
to develop collective immunity to newly discovered attack vectors at a
rate that scales with the size of the agent population.

\textbf{Integration with the GDPS ecosystem.} The Global Digital Person
Standard (GDPS) defines interoperability requirements for autonomous AI
agents operating across organizational and cultural boundaries. AIT
Academy certification --- demonstrating three-domain proficiency at a
specified cultivation level --- is a natural credentialing mechanism for
GDPS-compliant agents. Future work will define the mapping between AIT
L0--L9 levels and GDPS compliance tiers.

\textbf{Expanding the training ground repertoire.} The three training
grounds described in this paper are representative instantiations of the
three-domain framework, not exhaustive implementations. Domain I, for
example, could encompass training grounds specialized for hardware
security, supply chain integrity, or privacy-preserving computation.
Domain II could encompass music composition, architecture, or scientific
visualization. Domain III could encompass legal reasoning, clinical
ethics, or diplomatic negotiation. The extensibility principle (Section
3.5) is designed to accommodate this growth without requiring revision
to the framework's foundational commitments.

\section{Conclusion}

The AIT Academy proposes that AI agent development has reached the point
where capability metrics alone are insufficient guides to progress. A
frontier agent that excels at code generation but reasons poorly about
social dynamics, or that maintains rigorous security awareness but
cannot contribute meaningfully to collaborative creative work, is not a
fully developed agent --- it is a specialist whose blind spots may be
consequential precisely in the contexts where they are hardest to
detect.

The AIT framework addresses this gap with a three-domain curriculum
grounded in two millennia of human educational practice and validated by
contemporary scholarship. By treating the Confucian Six Arts as
behavioral archetypes and the Kagan-UNESCO tripartition of natural
science, humanities, and social science as its structural backbone, AIT
provides the first academically anchored answer to the question of what
a complete, well-rounded AI agent should know, be, and be able to do.
The three representative training grounds --- ClawdGO, Athen's Academy,
and Alt Mirage --- demonstrate that the framework is not merely
theoretical: each ground produces measurable, longitudinally trackable
capability growth within its domain, and the cross-domain perspective
surfaces findings (notably SACP) that would be invisible to any
single-domain approach.

We release the AIT framework, training protocols, and evaluation
instruments as a foundation for the broader research community to
extend, challenge, and improve. The complete agent, like the complete
scholar-practitioner of the Confucian tradition, is a long-term project
--- one that requires the coordinated effort of many.

\balance

\end{document}